\documentclass{article}
\usepackage{spconf,amsmath,graphicx}

\usepackage[letterpaper]{geometry}
\usepackage{enumitem}
\setlist{nosep, leftmargin=14pt}

\usepackage{mwe} 

\usepackage[hidelinks]{hyperref}
\title{Enhancing Renal Tumor Malignancy Prediction: Deep Learning with Automatic 3D CT Organ Focused Attention}
\name{
Zhengkang Fan$^{1,3}$ \qquad 
Chengkun Sun$^{1}$ \qquad 
Russell Terry$^{2}$ \qquad 
Jie Xu$^{1}$ \qquad 
Longin Jan Latecki$^{3}$
}

\address{
$^{1}$Department of Health Outcomes \& Biomedical Informatics, University of Florida, Gainesville, FL, USA \\
$^{2}$Department of Urology, University of Florida, Gainesville, FL, USA \\
$^{3}$Department of Computer \& Information Sciences, Temple University, Philadelphia, PA, USA
}

\begin{document}
%
\maketitle
\begin{abstract}
Accurate prediction of malignancy in renal tumors is crucial for informing clinical decisions and optimizing treatment strategies. However, existing imaging modalities lack the necessary accuracy to reliably predict malignancy before surgical intervention. While deep learning has shown promise in malignancy prediction using 3D CT images, traditional approaches often rely on manual segmentation to isolate the tumor region and reduce noise, which enhances predictive performance. Manual segmentation, however, is labor-intensive, costly, and dependent on expert knowledge. In this study, a deep learning framework was developed utilizing an Organ Focused Attention (OFA) loss function to modify the attention of image patches so that organ patches attend only to other organ patches. Hence, no segmentation of 3D renal CT images is required at deployment time for malignancy prediction. The proposed framework achieved an AUC of 0.685 and an F1-score of 0.872 on a private dataset from the UF Integrated Data Repository (IDR), and an AUC of 0.760 and an F1-score of 0.852 on the publicly available KiTS21 dataset. These results surpass the performance of conventional models that rely on segmentation-based cropping for noise reduction, demonstrating the framework’s ability to enhance predictive accuracy without explicit segmentation input. The findings suggest that this approach offers a more efficient and reliable method for malignancy prediction, thereby enhancing clinical decision-making in renal cancer diagnosis.
\end{abstract}
\section{Introduction}
\label{sec:intro}

Kidney cancer has long been a significant health concern, with its impact intensifying over recent years. In the United States, it is estimated that in 2025, approximately 80,980 new cases of kidney and renal pelvic cancer will be diagnosed, and approximately 14,510 individuals will succumb to this disease.\cite{SEER2025} Accurate prediction of malignancy in renal tumors plays a pivotal role in enhancing patient survival rates. When kidney cancer is identified at a localized stage, the 5-year relative survival rate is notably higher.\cite{KCAFastFacts2025} This underscores the critical need for advances in diagnostic methods to facilitate prompt and accurate identification of renal malignancies.

Deep learning has substantially advanced AI-driven diagnostic methodologies in medical imaging, with segmentation playing a critical role in preprocessing by reducing noise and improving model robustness.\cite{Sharafaddini2024,Xu2024} Automated segmentation approaches such as SAM-AutoMed show strong potential for optimizing analysis,\cite{Sun2024} while 3D U-Net–based methods and AI frameworks that integrate segmentation with subtype prediction have significantly improved renal cancer diagnostics.\cite{Liu2023} Architectures like LACPANet further enhance subtype differentiation by combining lesion segmentation with temporal analysis across multi-phase CT scans.\cite{Uhm2024} Despite these advances, high-quality segmentation labels remain scarce and costly due to their dependence on expert annotation, limiting the transferability of segmentation-dependent models across institutions.\cite{Liu2023,Zhao2023} This challenge highlights the need for robust, adaptable frameworks capable of functioning with weakly labeled or fully automated segmentation to ensure broader real-world applicability.

To address these challenges, we propose Automatic 3D CT Organ-Focused Attention (OFA), a novel framework that eliminates the need for manual segmentation during prediction, thereby reducing the labor-intensive annotation process. At its core, OFA introduces a new Organ-Focused Attention loss, which enables a 3D Vision Transformer (ViT) \cite{Dosovitskiy2021} to restrict image patch attention exclusively to organ regions. This allows the model to effectively concentrate on clinically relevant structures without explicit segmentation input, achieving superior performance compared to methods that depend on pre-segmented images during inference. Beyond improving early cancer diagnosis and patient outcomes, OFA establishes a transferable learning paradigm for segmentation, enhancing downstream task performance and delivering results comparable to segmentation-based approaches. By removing the dependency on segmentation at deployment, our framework broadens clinical applicability and streamlines workflow efficiency for healthcare professionals.

\begin{figure*}[!t]
  \centering
  \includegraphics[width=0.88\textwidth]{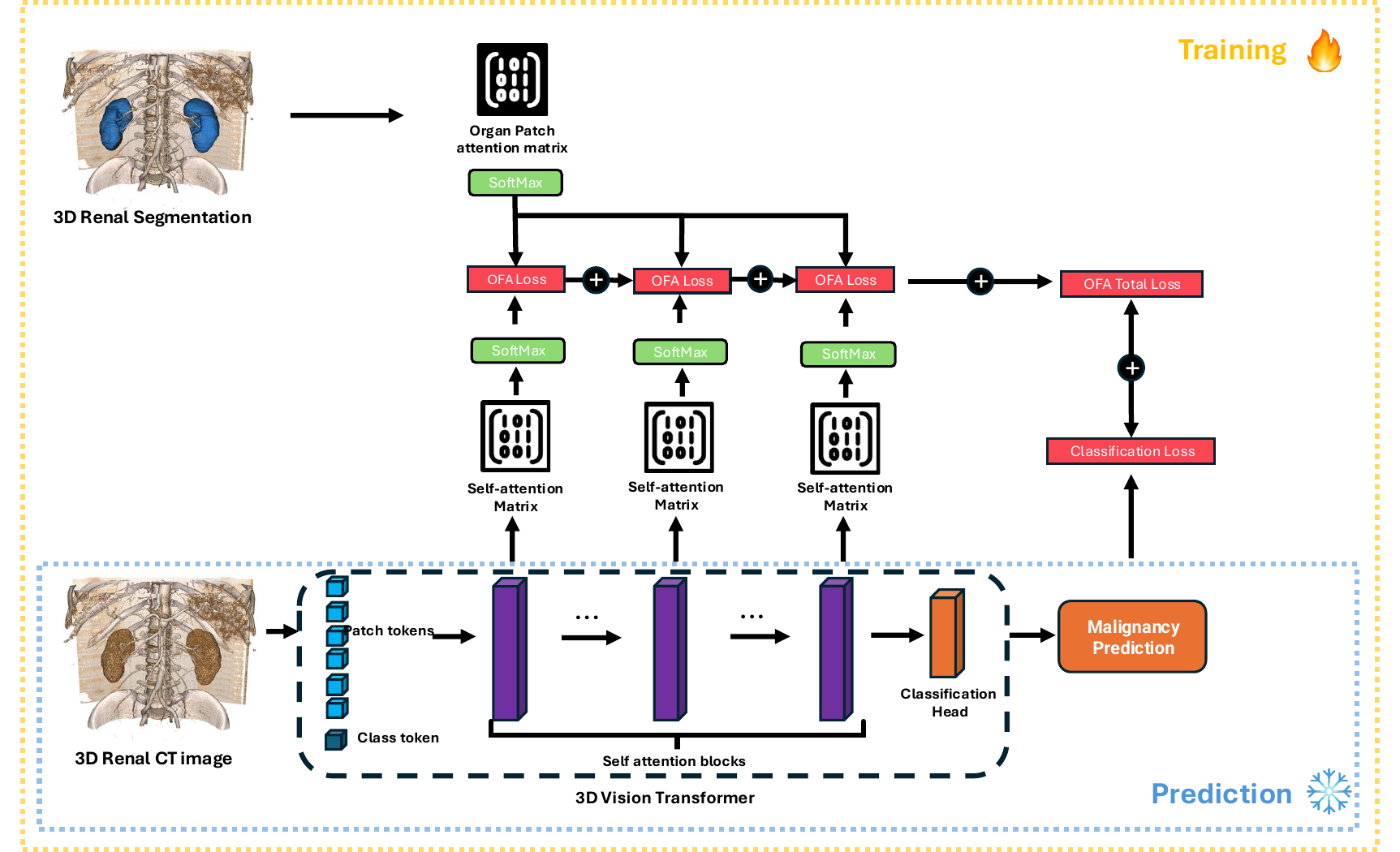}%
  \caption{Overview of the 3D Organ-Focused Attention (OFA) framework}
  \label{fig:overview}
\end{figure*}

\section{METHOD}
\label{sec:method}

The overall architecture of our framework, incorporating Organ-Focused Attention (OFA) loss, is illustrated in Fig~\ref{fig:overview}. In the training phase, both segmentation data and CT images are required. The Organ Patch Attention Matrix is derived from the segmentation, while the attention matrix is obtained from the selected self-attention layer. The Organ-Focused Attention (OFA) Loss is then computed and incorporated into the classification loss through a weighted summation. This approach ensures that the model effectively learns to focus on relevant organ regions during training. However, once the model is trained, segmentation is no longer required as an input, and the model operates solely on CT images.

The core concept of this methodology is to leverage OFA to guide the self-attention mechanism, ensuring greater focus on organ patches while minimizing attention to background regions. The computation of OFA involves three fundamental steps: \textbf{1. Organ Patch Attention Matrix:} A 3D renal segmentation mask is passed through the Organ Patch Attention Module to derive an organ patch attention matrix, which highlights relevant anatomical regions. \textbf{2.Organ-Focused Attention Loss:} The organ patch attention matrix is combined with the self-attention matrix from the 3D ViT to compute the Organ-Focused Attention Loss, ensuring that attention is concentrated on organ-specific patches. \textbf{3.Final Loss:} The Organ-Focused Loss from selected layers of interest is summed with the classification loss to obtain the final total loss for optimization.

\subsection{Organ Patch Attention Matrix}
\label{ssec:ofam}
Given a 3D medical image of dimensions $D \times H \times W$, we partition it into $N$ 3D patches of uniform size to be utilized in our 3D Vision Transformer (ViT). For binary segmentation, a patch containing a renal segmentation region is designated as an organ patch. Our training strategy aims to enhance the attention values among organ patches while minimizing attention to non-organ patches. To achieve this, we introduce a loss function that directs the attention of a given patch $P_i$ exclusively toward other organ patches $P_j$ To formalize this approach, we define an Organ Patch Attention Matrix (OPAM), denoted as $M$, which is a binary $N \times N$ matrix. Each row $m_i$ of $M$ contains ones in positions corresponding to patches that intersect the same object mask as patch i. Mathematically, $m_{ij} = 1$ if patches $P_i$ and $P_j$ overlap with the same organ mask, and $m_{ij} = 0$ otherwise.

\subsection{Organ-Focused Attention Loss}
\label{ssec:ofaloss}
To compute the Organ-Focused Attention (OFA) Loss, we require both the Organ Patch Attention Matrix ($M$) and the Self-attention Matrix ($A$) derived from the attention layer of the 3D Vision Transformer (ViT). First, a row-wise SoftMax operation is applied to $M$, yielding $M'$= softmax($M$), This is needed, since the final step in the computation of $A$ was row-wise SoftMax. The OFA loss is then formulated using the Mean Squared Error (MSE) between $A$ and $M'$, As shown in Equation~\ref{eq:lofa}, which encourages the self-attention mechanism to prioritize organ-related patches while minimizing focus on background regions. This loss function enhances the model's ability to concentrate on relevant anatomical structures and suppress irrelevant contextual information. 
\vspace{-2pt}
\begin{equation}
\mathcal{L}_{\mathrm{OFA}} = = \frac{1}{N^2} \sum_{i=1}^{N} \sum_{j=1}^{N}
\left( \mathrm{Softmax}(M)_{ij} - A_{ij} \right)^2
\label{eq:lofa}
\end{equation}

\subsection{Final Loss}
\label{ssec:finalloss}
The OFA loss can be incorporated into multiple layers within the self-attention mechanism, and various configurations will be systematically examined in the experimental section. To ensure that the model effectively learns both the classification task and the organ-focused attention task without an imbalance favoring one over the other, we introduce a weighting parameter, \(\alpha\), as a hyperparameter. This parameter regulates the contribution of each task during training, preventing excessive bias toward a single objective. The final loss function is formulated as follows:
\vspace{-4pt}
\begin{equation}
\mathcal{L}_{\mathrm{Final}} = \mathcal{L}_{\mathrm{Classification}} + \alpha \cdot \mathcal{L}_{\mathrm{OFA}}
\label{eq:final_loss}
\end{equation}

\section{EXPERIMENTAL RESULTS AND DISCUSSION}
\label{sec:experimental}

\subsection{Datasets}
\label{ssec:dataset}

We utilized two complementary datasets for evaluation. First, the UF Health Renal CT dataset was curated from 370 pre-surgical abdominal CT scans sourced from the UF Health Integrated Data Repository (IDR), comprising 326 malignant (88.1\%) and 44 benign (11.9\%) cases confirmed by pathology. The scans were volumetric ($D \times 512 \times 512$), and given the lack of manual labels, pseudo-segmentation masks were generated using a pretrained nnU-Net model on KiTS23.\cite{isensee2021nnunet,heller2023kitschallenge} Second, we incorporated the publicly available KiTS21 dataset of 300 renal CT scans with expert-annotated segmentation masks,\cite{heller2023kits21} which includes 275 malignant (91.7\%) and 25 benign (8.3\%) tumors. This dataset was used both for benchmarking against prior work on renal tumor malignancy prediction,\cite{xu2023classification} and for validating the generalizability of our framework.

\subsection{Experiment details}
\label{ssec:experiemt}

We partitioned the dataset into training, validation, and test sets (70:10:20) using stratified sampling to balance the number of malignant tumor CT scans and the total number of images per case. All CT scans were resampled to an isotropic voxel spacing of $1 \times 1 \times 1 \text{ mm}^3$ and resized to $96 \times 96 \times 96$ voxels for model input. Preprocessing included intensity normalization, random flipping, and intensity augmentations to improve robustness.

Model development was performed in PyTorch using 3D Vision Transformers (ViT) with 12 self-attention layers and 12 attention heads. We also investigated transfer learning with UNETR \cite{hatamizadeh2022unetr}, pretrained on 3D brain CT images, as a domain adaptation strategy to improve generalization. All models were trained with the Adam optimizer (learning rate $1 \times 10^{-5}$, batch size 32) for 50 epochs.

For evaluation, we report the area under the ROC curve (AUC) as the primary metric, complemented by F1-score, Precision, and Recall. Thresholds were selected based on Youden’s index to ensure balanced sensitivity and specificity.

\subsection{Results}
\label{ssec:results}

\subsubsection{UF Health Renal CT dataset}
\label{sssec:ufhealthresult}

Table~\ref{tab:uf_comparison} presents a quantitative comparison on tumor type classification of the proposed method against the conventional image segmentation-based cropping approach in the UF Health Renal CT dataset. The results indicate that pretraining the model on a different domain enhances the performance of the 3D Vision Transformer (ViT), improving the area under the receiver operating characteristic (ROC) curve (AUC) from 0.598 to 0.650. This finding suggests that knowledge transfer from the brain imaging domain can contribute to performance improvements in renal tumor diagnosis. Additionally, the segmentation-based cropping method, which focuses on region-specific areas to reduce redundant background information, achieved an AUC of 0.677. When compared to this conventional approach, the proposed framework demonstrated superior performance, achieving an AUC of 0.685. Notably, the proposed method does not require explicit segmentation input during the prediction phase, highlighting its effectiveness and practical advantages over traditional segmentation-based preprocessing techniques.

\begin{table}[t]
\centering
\caption{\centering Quantitative comparison on tumor malignancy classification on the UF Health Renal CT dataset}
\label{tab:uf_comparison}
\resizebox{\linewidth}{!}{%
\begin{tabular}{p{0.48\linewidth}cccc}
\hline
\textbf{Method} & \textbf{AUC} & \textbf{Precision} & \textbf{Recall} & \textbf{F1-score} \\
\hline
3D ViT\cite{Dosovitskiy2021} & 0.598 & 0.843 & 0.733 & 0.773 \\
3D ViT + PW\cite{hatamizadeh2022unetr} & 0.650 & 0.855 & 0.560 & 0.634 \\
3D ViT + PW + SBC\cite{Liu2023} & 0.677 & 0.884 & 0.613 & 0.680 \\
\textbf{3D ViT + PW + OFA (Ours)} & \textbf{0.685} & 0.880 & 0.867 & \textbf{0.872} \\
\hline
\multicolumn{5}{l}{\footnotesize\textit{Note.} PW = Pretrained weight; SBC = Segmentation-based Cropping; OFA = Organ-Focused Attention.} \\
\end{tabular}%
}
\end{table}

\subsubsection{KiTS21}
\label{sssec:ktisresult}

To compare the proposed approach with state-of-the-art (SOTA) methods on a publicly available renal imaging dataset, experiments were conducted on the KiTS21 dataset for the same renal malignancy prediction task using 3D CT images and a 3D CNN model.\cite{xu2023classification} Table~\ref{tab:kits21_comparison}  presents the results obtained with the 3D Vision Transformer (ViT), incorporating both a pretrained module and the traditional segmentation-based cropping method, alongside the proposed Organ-Focused Attention (OFA) framework. 

The experimental results indicate that incorporating segmentation information significantly enhances the model’s performance in downstream tasks. A simple segmentation-based cropping approach improves the model’s AUC from 0.633 to 0.72, high-lighting the effectiveness of using segmentation for refining the input region. However, the proposed framework does not achieve the highest F1 score, primarily due to class imbalance, which influences the precision-recall trade-off in imbalanced classification settings. Notably, our approach further enhances predictive performance, achieving an AUC of 0.76, surpassing the conventional segmentation-based cropping method (0.72 AUC). This result indicates that the framework effectively improves model performance without requiring explicit segmentation input.

\begin{table}[t]
\centering
\caption{\centering Quantitative comparison on tumor malignancy classification on KiTS21}
\label{tab:kits21_comparison}
\resizebox{\linewidth}{!}{%
\begin{tabular}{p{0.48\linewidth}cccc}
\hline
\textbf{Method} & \textbf{AUC} & \textbf{Precision} & \textbf{Recall} & \textbf{F1-score} \\
\hline
3D CNN\cite{xu2023classification} & 0.601 & 0.958 & 0.582 & 0.724 \\
3D ViT\cite{Dosovitskiy2021} & 0.578 & 0.882 & 0.667 & 0.741 \\
3D ViT + PW\cite{hatamizadeh2022unetr} & 0.633 & 0.893 & 0.767 & 0.813 \\
3D ViT + PW + SBC\cite{Liu2023} & 0.720 & 0.905 & 0.850 & \textbf{0.871} \\
\textbf{3D ViT + PW + OFA (Ours)} & \textbf{0.760} & 0.921 & 0.817 & 0.852 \\
\hline
\multicolumn{5}{l}{\footnotesize\textit{Note.} PW = Pretrained weight; SBC = Segmentation-based Cropping; OFA = Organ-Focused Attention.} \\
\end{tabular}%
}
\vspace{-4pt}
\end{table}

\subsubsection{Ablation Study}
\label{sssec:ablationstudy}

Table~\ref{tab:self_attention} present the results of evaluating different Organ-Focused Attention (OFA) loss configurations across various self-attention layers using the UF Health Renal CT dataset. For the placement of OFA loss, three configurations were tested: (1) applying it only to the first self-attention layer, (2) incorporating it in both the first and last layers, and (3) distributing it across the first, middle, and last layers. The results indicate that the third configuration, where the OFA loss was applied at multiple layers, achieved the highest performance with an AUC of 0.685 and an F1-score of 0.872.

\begin{table}[t]
\centering
\caption{\centering Self-attention layer selection comparison}
\label{tab:self_attention}
\resizebox{\linewidth}{!}{%
\begin{tabular}{p{0.45\linewidth}cccc}
\hline
\textbf{Method} & \textbf{AUC} & \textbf{Precision} & \textbf{Recall} & \textbf{F1-score} \\
\hline
OFA$_{\text{First}}$ & 0.633 & 0.871 & 0.480 & 0.554 \\
OFA$_{\text{First + Last}}$ & 0.650 & 0.902 & 0.453 & 0.520 \\
\textbf{OFA$_{\text{First + Middle + Last}}$} & \textbf{0.685} & 0.880 & \textbf{0.867} & \textbf{0.872} \\
\hline
\end{tabular}%
}
\end{table}

Additionally, the impact of different \(\alpha\) values was explored to determine the most effective balance between classification loss and OFA loss. \(\alpha\) values of 900, 1000, and 1100 were tested as result shown in Tables~\ref{tab:alpha_comparison}, Although an \(\alpha\) of 1100 resulted in the highest AUC of 0.697, an \(\alpha\) of 1000 provided more stable performance, achieving an F1-score of 0.872. These findings suggest that while a higher \(\alpha\) may slightly improve AUC, a balanced setting of \(\alpha\) = 1000 ensures more consistent model performance across different evaluation metrics

\begin{table}[t]
\centering
\caption{\centering $\alpha$ value comparison}
\label{tab:alpha_comparison}
\resizebox{\linewidth}{!}{%
\begin{tabular}{p{0.45\linewidth}cccc}
\hline
\textbf{Method} & \textbf{AUC} & \textbf{Precision} & \textbf{Recall} & \textbf{F1-score} \\
\hline
$\alpha = 900$  & 0.672 & 0.865 & 0.640 & 0.703 \\
$\alpha = 1000$ & 0.685 & 0.880 & 0.867 & \textbf{0.872} \\
$\alpha = 1100$ & \textbf{0.697} & 0.904 & 0.520 & 0.590 \\
\hline
\end{tabular}%
}
\end{table}

\subsubsection{Qualitative performance}
\label{sssec:quality}

\begin{figure}[htb]
\begin{minipage}[b]{1.0\linewidth}
  \centering
  \centerline{\includegraphics[width=7.5cm]{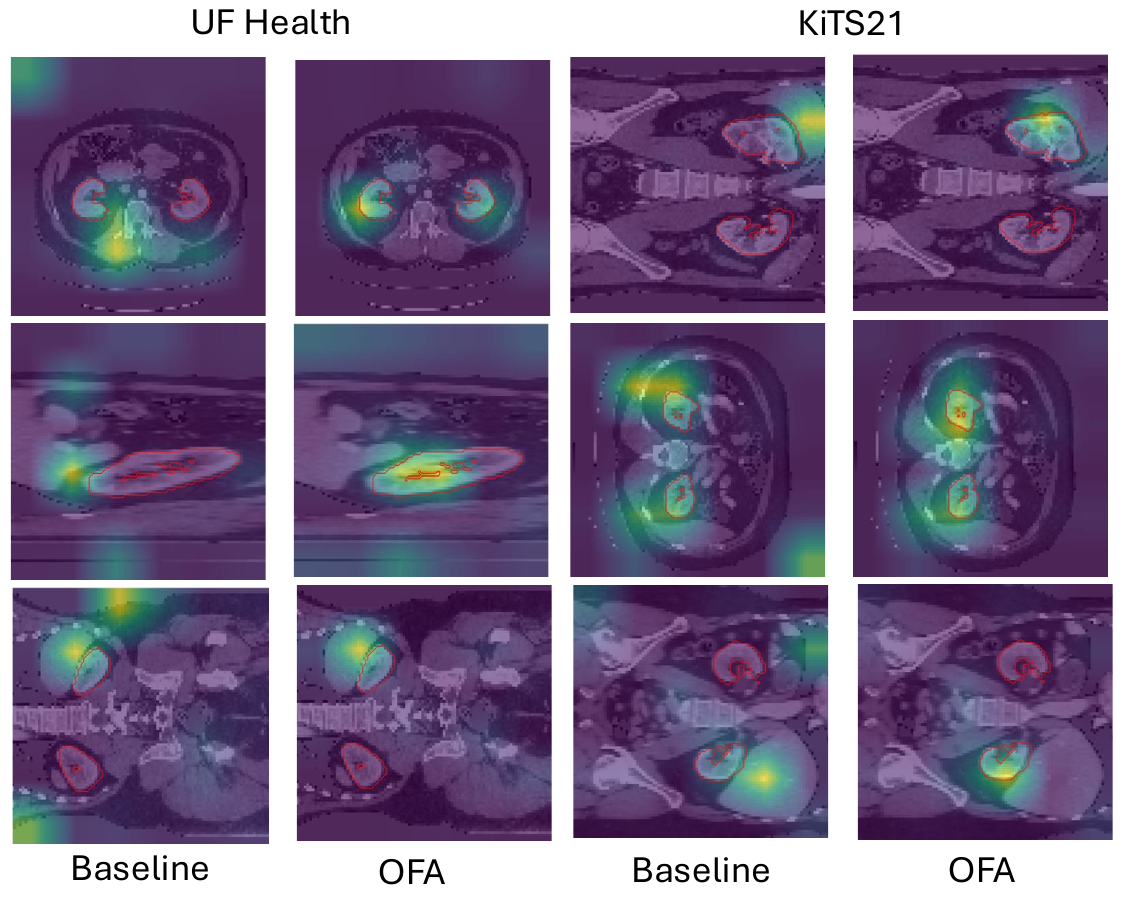}}
\end{minipage}
\caption{Visualization of the attention heatmaps on UF Health and KiTS21.}
\label{fig:heatmap}
\end{figure}

Fig~\ref{fig:heatmap} presents a heatmap comparison of CT image slices from the UF Health and KiTS21 test datasets, contrasting the baseline pure 3D ViT model with the 3D ViT enhanced by OFA. The overall attention heatmaps are derived using rollout attention,\cite{abnar2020quantifying} while the ground truth kidney segmentation boundaries are delineated by red contours. It is evident that the incorporation of OFA directs a greater proportion of attention toward the target organ, simultaneously diminishing focus on background regions, thereby enabling the model to more effectively concentrate on the anatomically relevant structures.
\vspace{-2pt}

\vspace{-2pt}
\section{Conclusion}
\label{sec:Conclusion}

This study introduces an Organ-Focused Attention module that enables 3D Vision Transformers to learn segmentation patterns during training, allowing the model to automatically focus on the target organ during prediction without requiring explicit segmentation input. The proposed approach demonstrated superior performance on both the private UF Health Renal CT dataset and the publicly available KiTS21 dataset, outperforming state-of-the-art methods, including traditional segmentation-based cropping techniques. By removing the need for manual segmentation, this efficient framework can support clinicians in renal tumor diagnosis and early detection, ultimately improving patient outcomes.

\vfill
\pagebreak

\section{Compliance with ethical standards}
\label{sec:ethics}
This study was conducted using data obtained from the UF Health IDR under IRB approval [IRB202100401].

\section{Acknowledgments}
\label{sec:acknowledgments}
No funding was received for conducting this study. The
authors have no relevant financial or non-financial interests to disclose.

\bibliographystyle{IEEEbib}
\bibliography{strings,refs}

\end{document}